# Pioneer dataset and automatic recognition of Urdu handwritten characters using a deep autoencoder and convolutional neural network

Hazrat Ali[1,*], Ahsan Ullah[1], Talha Iqbal[2], Shahid Khattak[1]

[1]Department of Electrical and Computer Engineering, COMSATS University Islamabad, Abbottabad Campus, Abbottabad, Pakistan.
[2]Lambe Institute of Translational Research, National University of Ireland, Galway, Ireland.
hazratali@cuiatd.edu.pk, engr.ahsan86@gmail.com,
t.iqbal1@nulgalway.ie, skhattak@cuiatd.edu.pk
*Correspondence: hazratali@cuiatd.edu.pk

**Abstract.** Automatic recognition of Urdu handwritten digits and characters, is a challenging task. It has applications in postal address reading, bank's cheque processing, and digitization and preservation of handwritten manuscripts from old ages. While there exists a significant work for automatic recognition of handwritten English characters and other major languages of the world, the work done for Urdu language is extremely insufficient. This paper has two goals. Firstly, we introduce a pioneer dataset for handwritten digits and characters of Urdu, containing samples from more than 900 individuals. Secondly, we report results for automatic recognition of handwritten digits and characters as achieved by using deep auto-encoder network and convolutional neural network. More specifically, we use a two-layer and a three-layer deep autoencoder network and convolutional neural network and evaluate the two frameworks in terms of recognition accuracy. The proposed framework of deep autoencoder can successfully recognize digits and characters with an accuracy of 97% for digits only, 81% for characters only and 82% for both digits and characters simultaneously. In comparison, the framework of convolutional neural network has accuracy of 96.7% for digits only, 86.5% for characters only and 82.7% for both digits and characters simultaneously. These frameworks can serve as baselines for future research on Urdu handwritten text.

# 1    Introduction

Handwritten text recognition is an interesting task due to its tremendous applications such as to convert handwritten documents into a digital format, reading house numbers automatically, postal address reading and robotics [1], [2], [32], [33], [34]. Unlike a typical text in one single font, handwritten text recognition is challenging due to the fact that writing styles vary from person to person.

The Urdu language carries extreme importance as one of the largest languages of the world and the national language of Pakistan. Urdu text shares similarities with Arabic and Persian text. This work presents a framework for automatic recognition of Urdu handwritten letters. The task is less explored for Urdu. One primary reason that there has been no dataset available for Urdu handwritten text. To address this, we introduce a new dataset of Urdu handwritten digits and characters. The motivation comes from the fact that a standard dataset of Urdu handwritten text does not exist, which may serve as a baseline for research work. Urdu is one of the largest languages of the world, being the first language of more than 60 million people (and more than 329 million people if combined with Hindi as the two languages are greatly the same in spoken form). Unfortunately, there seems to be very less or no work on Urdu language processing mainly due to unavailability of language resource. Besides, a standard dataset would help out the research community as unlike English and many other languages, Urdu text recognition is more challenging due to the presence of diacritics. Similar (but not the same) diacritics are found in Arabic and Persian languages, and thus, any research development on Urdu text recognition would eventually ease out progress in research work on handwritten text recognition of many more languages. While there has been the UCOM dataset [31] reported for Urdu text, several differences exist between the UCOM dataset and our dataset. Firstly, the UCOM offline dataset has been developed for continuous text of Urdu. Our dataset is for isolated characters of Urdu hand-written text. Secondly, the UCOM dataset, as described by the authors in [31], contains text for 600 pages of Urdu text and the number of different individuals who have written the text is limited to 100, while our dataset contains text from 900 individuals. Thirdly, The UCOM dataset contains text in Nasta'liq style only while our dataset contains hand-written samples in different styles and variations, thus covering a more diverse range of writing (font) styles.

Deep learning (a sub branch of machine learning) algorithms have been popular for automatic recognition of digits and characters of different languages. Deep networks can be trained in supervised fashion requiring labels, or in an unsupervised way without requirements of labels [3], [4], [5]. In this work, we use an autoencoder network and a convolutional neural network (CNN) trained with 85% portion of the dataset and tested with the remaining 15% of the data. Moreover, these models are evaluated for configuration with two hidden layers and three hidden layers.

The rest of the paper is organized as follows. Section 2 provides literature review on existing work done for Urdu text recognition. In Section 3, we describe the dataset developed, source of the data, pre-processing and segmentation steps. We describe the use of a deep autoencoder network and CNN in Section 4. Results are presented in Section 5 and finally; the paper is concluded in Section 6.

## 2       Literature Review

For character recognition, machine learning techniques such as deep neural network and CNN have been used. Arnold et al., used neural networks for character recognition [6]. Similarly in [7], [8], CNN has been used for Chinese characters recognition. A stacked denoising autoencoder has been used in [9] for offline Urdu character recognition. However, the work in [9] is limited to optical character recognition of Nastaliq fonts only. Hussain et al., proposed an offline OCR system to recognize only eight Arabic handwritten characters with accuracy rate of 77.25% [10]. The framework proposed by Elenwar et al., [11] used Arabic characters database containing 1814 characters for training and 435 characters for testing. The database used in [12] is prepared by only four writers leading to low generalization. A database for Arabic characters is presented in [13] in which the authors performed pre-processing steps to avoid noise in the printed database. Another database for Arabic characters consists of 28 thousand characters of Arabic language written by 100 different writers [14]. A similar work has been reported by [14] as they target online recognition of Urdu characters collected from 100 writers for recognition of seven characters only. This review shows that most of the work done in the field of Urdu character recognition is for small datasets and with very limited generalization capability. Some progresses on Urdu script recognition are also presented in [13] and [14], but those are for printed text (typically popular with OCR applications) while we are developing an algorithm for handwritten Urdu text recognition.

To the best of our knowledge, there is no dataset available for Urdu handwritten digits and characters. We present a new dataset consisting of handwritten digits and characters of Urdu, written by 900 different individuals. This dataset goes through different pre-processing stages like RGB to grey-scale conversion, noise removal and segmentation. Furthermore, we use an unsupervised algorithm called autoencoder and CNN for recognition of Urdu handwritten characters, a task not explored before to the best of our knowledge.

## 3       Data collection and pre-processing stage

### 3.1       Data Acquisition

We collected the data from 900+ individuals of different age group following a tabular format as shown in Fig. 1. After data cleaning and pre-processing, we retain the data of 900 individuals while discarding the remaining samples due to inconsistency in writing quality or missing entries. The individuals belong to different age groups in the range of 22 to 60 years. The writers are a mix of native and non—native speakers of Urdu, however, this has no direct impact on the writing style. Hence, this factor has not been considered in selection of writers. The dataset contains samples for ten digits and 40 characters, as can be seen in Fig. 1. As expected, the writing style differs for different individuals, thus introducing diverse writing style into the dataset. The writing samples were then scanned and stored in a computer[1].

| ۹ | ۸ | ۷ | ۶ | ۵ | ۴ | ۳ | ۲ | ۱ | ٬ |
| ح | خج | ج | ث | ﭦ | ﺚ | پ | ب | آ | ا |
| ﭧ | س | ژ | ز | ڗ | ﺮ | ڈ | د | خ |
| گ | ﯼ | ق | ﺖ | ﻉ | ظ | ط | ض | ص |
| ﻊ | ﻯ | ﮮ | ﻪ | و | ن | (ن) | م | ل |

**Fig. 1.** Sample of dataset of Urdu (digits and characters)

### 3.2    Pre-processing of dataset

Pre-processing of dataset included scanning of dataset of Urdu digits and characters written by different writers, removing noise by thresholding, segmentation and finally compression of each image to $28 \times 28$ pixels.

### 3.3    Segmentation

In segmentation process, the group of digits and characters are segmented into individual digits and characters as shown in Fig. 2.

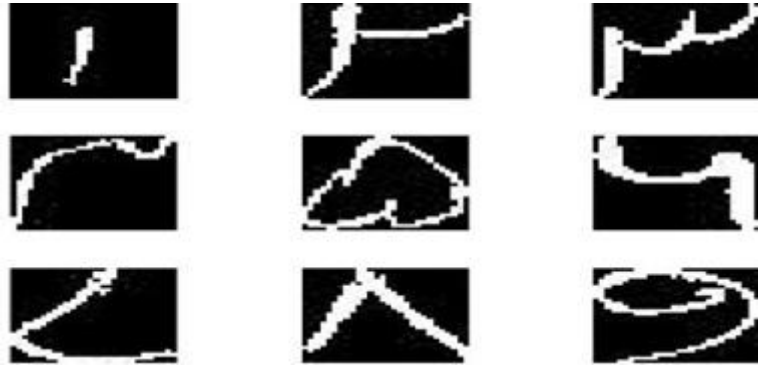

**Fig. 2.** Segmentation of dataset

---

In our case we have 900 samples of images and each image consists of 10 digits and 40 characters. We divide each sample of image into 50 small images of size $28 \times 28$ pixels thus saving each individual digit and character as a separate image. After segmentation process and removal of selected noisy samples, we get a total of 45 thousand individual images. Out of these, the training and test sets are selected randomly with a ratio of 85% and 15% respectively. To avoid any bias in the training model, the training/test split is subject-independent as none of the samples in training and test sets is from the same individual. This fulfills the requirements of completely independent training and test sets.

## 4    Proposed model

The proposed framework is based on a deep autoencoder network and a convolutional neural network. Specifically, we train two-layer and three-layer networks and compare the results.

### 4.1    Deep autoencoders

An autoencoder neural network is an unsupervised learning algorithm that uses back-propagation to set the target value to be equal to the inputs [20], [21]. An autoencoder model is popular for its ability to learn important features by reconstructing the input at the output. During the reconstruction process, the autoencoder tries to learn useful representations of the (raw) input. An autoencoder consists of three or more layers: an input layer; some number of hidden layers, which form the encoding; and an output layer, whose dimension is the same as input layer. In order to get useful representation of input, the number of neurons in the hidden layer is kept smaller than the input. For example, if the input has 784 neurons, the number of neurons in the next layer is less than 784 to get a compressed representation of the input. By compressing the input, the auto encoder tends to learn the best representation (features) from the input from which the input can be reconstructed easily and efficiently at the output [15], [16], [20], [21]. The hidden layer and the output layer perform the important tasks in autoencoder, as the hidden layer encodes the input, and the output layer decodes it to get the original form of the input data. Moreover, another good thing about autoencoder is that the hidden layer can be configured to reduce the dimension or size of the input. This characteristic of an autoencoder is one of the many ways to learn useful features of the input data as shown in Figure 3.

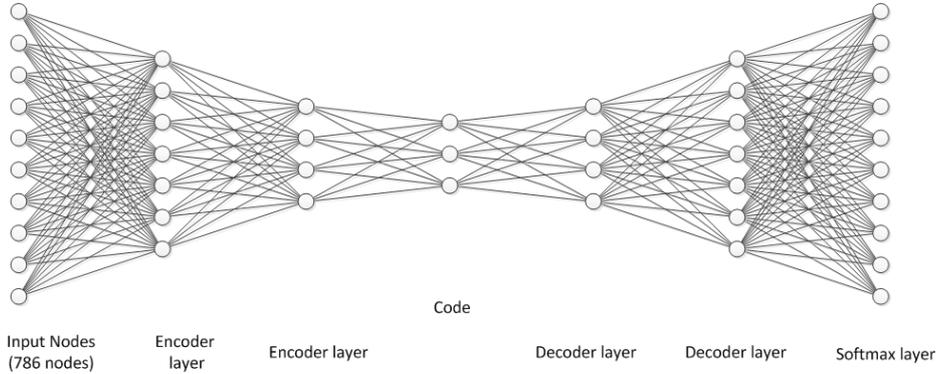

Input Nodes (786 nodes) | Encoder layer | Encoder layer | Code | Decoder layer | Decoder layer | Softmax layer

A two layer auto-encoder architecture. Number of nodes shown in the figure are arbitrary for demonstration purpose only. The input layer has 786 units for an image of 28 by 28. The final layer is a softmax layer for classification. The number of units in the soft max layer is equal to the distinct characters

**Fig. 3.** A basic two layer autoencoder architecture. For our model, the number of nodes in the final layer is equal to the number of distinct characters. After experimentation, we choose 100 nodes in the first hidden layer and 100 nodes in the second hidden layer. These nodes are fully connected.

The standard way to train an autoencoder is to use back propagation to reduce the reconstruction error, but it is generally very difficult to optimize non-linear autoencoders with multiple hidden layers having hundreds of thousands of parameters. That is why autoencoder is trained in a greedy layer wise manner, i.e., we train one layer at a time, which can find a good set of parameters quickly, even in deep networks with millions of parameters and many hidden layers [17]. Autoencoder is extensively used in different areas of solving deep learning problems in an efficient way. Denoising autoencoder is used to reconstruct corrupted data of input in order to get good efficiency [18]. Sparse autoencoder approach is used to automatically learn features from unlabeled data [19]. Examples of similar work for recognizing digits appear on *reading house numbers from street level photos* in [22].

In our case, we use autoencoders with two and three hidden layers respectively with different numbers of neurons in each layer to gain high accuracy. Furthermore, we use the scaled conjugate gradient back-propagation algorithm, which is a network training function to update weight and bias values. Further, we use L2 weight regularization to control the influence of regularization. The autoenoder mainly serves as an unsupervised model. However, at the final layer of the model, we use the softmax classification which turns the overall model into a semi-supervised learning model, an approach most common for classification tasks. The hyper-parameters are optimized by using a grid search approach and training several models to identify the best choice for our task.

Overall training process of an autoencoder consists of the following steps:

- Pre-training step: Autoencoders are trained in a greedy manner, one layer at a time using unsupervised data.

- Softmax layer/last layer: Supervised data is used to train the last layer.
- Fine-tuning: To fine tune, we use back-propagation using supervised data to get the optimal accuracy rate.

The hyper-parameters are optimized by using a grid search approach and training several models to identify the best choice for our task.

### 4.2    Convolutional neural network

CNN is inspired from biological processes [23]. The connectivity pattern of the neurons resembles the organization of our visual cortex. CNN requires less pre-processing as compared to other image classification algorithms as it uses variation of multilayer perceptions [24]. CNN has become popular in many applications such as in recommender system, image and video recognition and natural language processing [25], [26], [27]. Like an ordinary neural network, neurons having some learnable weights and biases are the basic building blocks of a CNN. Each neuron performs dot product on the input given to it. Then a non-linearity is applied to the output of the dot product [26]. Overall, a convolution layer performs convolution using image patches and uses kernels with learnable weights. The output of a convolution layer becomes the input of a pooling layer and this goes on depending upon the number of layers in the network. The last layer of CNN is fully connected layer and performs the classification task. The whole network gives us a probability score for each class at the output. CNN is different from a simple neural network in a sense that it makes explicit assumption about the input i.e., input is always an image. With this assumption, we are able to encode some certain properties into our architecture, which in turn helps us to reduce amount of parameters in the neural network. Low level convolutional layers extracts low level information from the image like edges and corners while as we move further deeper into the network, convolution layer extracts high level information such as a complete character or number.

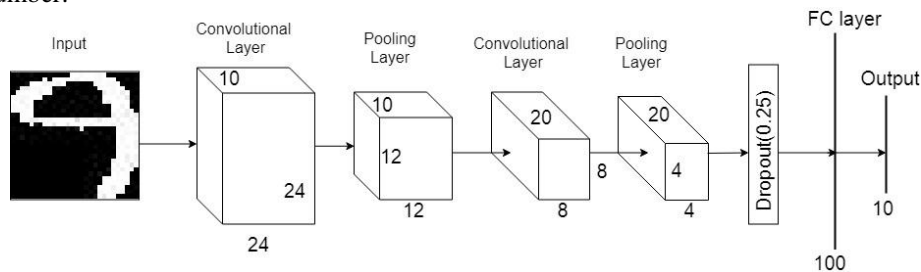

**Fig. 4.** Basic Convolutional Neural Network Framework.

A simple CNN can be thought of as a sequence of layers and each layer uses functions differentiable almost everywhere, to transform output of one layer to another [27], [28], [29], [30]. Main building layers are: Convolutional layer (CONV) with activation function (ReLU), pooling layer (POOL) and fully-connected (FC) layer. We stack these layers to make architecture of convolutional neural network, as in Figure 4. Each of these layers is elaborated below:

- Input holds image raw pixel values in shape of width, height and color channels.
- CONV layer computes dot product of weights and a small connected region of the input volume to give us output of neurons.
- ReLU is activation function that is applied element-wise. ReLU function performs thresholding at zero i.e., given input $x$, it selects $max(0, x)$.
- POOL layer is used for down-sampling the spatial dimensions i.e. width and height, so that we have reduced volume for further processing.
- FC layer is the last layer and it computes each class probability score. In this layer, as the name implies, every neuron in each layer is connected to all the neurons in previous layer.

In this way, the CNN transforms the input image layer by layer using pixel values to final class probability scores. CNN is independent from the use of prior knowledge and human effort in feature designing, which is its major advantage.

## 5      Simulation results

The available dataset is divided into training and test set with a ratio of 85% and 15% respectively. All the experiments were performed on a core i5 CPU with 3 GHz processor capacity. Experiments were carried out for different combinations of the hyper-parameters of the autoencoder and convolutional neural network (such as the number of neurons, number of hidden layers, and size of hidden layers). More specifically, we run the experiments for three different settings:

- Digits from ١ to ٩.
- Characters from Alif (ا) to Yaa (ے)
- Both digits and characters from 1 (١) to Yaa (ے).

We obtain the experiment for these three settings and discuss the results one by one.

### 5.1      Training an autoencoder

The autoencoder model is trained for two and three hidden layers and the result for each digit is shown in appendix I and appendix II. Result for each character is shown in appendix III and appendix IV. Appendix V and appendix VI show the results when autoencoder is trained on overall dataset i.e. on digits as well as characters using 2 and 3 hidden layers, respectively.

Most importantly, from the results of individual characters, it can be noticed that the accuracy rate is higher for those characters which have no similarity with other characters. It is found that similarities between characters like alif (ا) and digit 1 reduce the accuracy rate i.e., for digit 1, accuracy rate is 61.8% and for alif, accuracy rate is 51.8% as shown in appendix V.

Table 1 shows the training parameters used in our model training in order to get optimal results. It includes the number of neurons in each layer of an autoencoders, time taken for each layer for training process, number of iterations and learning rate.

Figure 5 through Figure 8 show some of the training performances in which error rate is a function of number of iterations. After analysis, it is found that as we increase the number of iterations and number of hidden layers, the error rate decreases.

Table 5 shows compendium of the results for 2 and 3 hidden layers. It is obvious that addition of third layer has resulted in increasing the accuracy.

**Table 1.** Training parameters for deep autoencoder (All experiments performed on a core i5 CPU with 3 GHz processor)

| Training type | 1st Layer Iterations = 350, Learning rate = 0.15 | | 2nd layer Iterations = 300, Learning rate = 0.1 | | 3rd layer Iterations = 350, Learning rate = 0.1 | |
|---|---|---|---|---|---|---|
| | Neurons | Time | Neurons | Time | Neurons | Time |
| Digits with 2 Hidden layers | 100 | 6:04 | 50 | 0:37 | - | - |
| Digits with 3 Hidden layers | 100 | 6:45 | 100 | 0:41 | 50 | 0:39 |
| Characters with 2 Hidden layers | 100 | 21:32 | 50 | 2:25 | - | - |
| Characters with 3 Hidden layers | 100 | 21:32 | 100 | 3:30 | 50 | 2:36 |
| Digits and Characters with 2 Hidden layers | 100 | 32:08 | 50 | 2:45 | - | - |
| Digits and Characters with 3 Hidden layers | 100 | 32:08 | 100 | 3:35 | 50 | 2:30 |

**Table 2.** Training parameters for Convolutional Neural Network

| Training type | 1st layer. Iterations = 20, | 2nd layer. Iterations = 20, | 3rd layer. Iterations = 20, |
|---|---|---|---|

| | Learning rate = 0.15 | Learning rate = 0.1 | Learning rate = 0.1 |
|---|---|---|---|
| | Neurons | Neurons | Neurons |
| Digits with 2 Hidden layers | 100 | 50 | - |
| Digits with 3 Hidden layers | 100 | 100 | 50 |
| Characters with 2 Hidden layers | 100 | 50 | - |
| Characters with 3 Hidden layers | 100 | 100 | 50 |
| Digits and Characters with 2 Hidden layers | 100 | 50 | - |
| Digits and Characters with 3 Hidden layers | 100 | 100 | 50 |

**Table 3.** Optimal parameters used in simulation for autoencoder and CNN

| Layer number | L2 weight Regularization | | Sparsity Regularization | | Sparsity | |
|---|---|---|---|---|---|---|
| | Autoen-coder | CNN | Auto-en-coder | CNN | Auto-en-coder | CNN |
| **1** | 0.004 | 0.09 | 4 | 6 | 0.15 | 0.1 |
| **2** | 0.002 | 0.06 | 4 | 6 | 0.1 | 0.3 |
| **3** | 0.002 | 0.06 | 4 | 6 | 0.1 | 0.3 |

**Table 4.** Summary of Results for autoencoder

| Type of autoencoder | Accuracy (%) | | |
|---|---|---|---|
| | Characters | Digits | Overall |
| **2 layered** | 77.6 | 96.8 | 80 |
| **3 layered** | 81.2 | 97.3 | 82 |

## 5.2    Training convolution neural network

We kept all the parameters same as those used for auto-encoder except the number of epochs for CNN i.e. 20 epochs (see Table 2). The optimal parameters are chosen through empirical results and are shown in Table 4. The optimal parameters are chosen through empirical results and are shown in Table 2.The convolution neural network model is trained for two, three and four hidden layers. The results are reported in Table

5. Loss function is categorical cross-entropy, batch size is 16 while number of epochs are kept 20. The optimizer for this model is **rmsprop**. It is obvious from the results that adding a third layer improves the overall accuracy but note that by adding fourth layer accuracy for digits and characters decreases while overall accuracy remains same.

**Table 5.** Summary of Results for auto-encoder and CNN models

| Training type | Accuracy (2 hidden layers) | | Error (2 hidden layers) | | Accuracy (3 hidden layers) | | Error rate (3 hidden layers) | |
|---|---|---|---|---|---|---|---|---|
| | Auto encoder | CNN | Auto encoder | CNN | Auto encoder | CNN | Auto encoder | CNN |
| Digits (1 – 9) | 96.8% | 95.6% | 3.2% | 4.4% | 97.3% | 96.7% | 2.7% | 3.3% |
| Characters | 77.6% | 69.4% | 22.4% | 30.6% | 81.2% | 86.6% | 18.8% | 13.4% |
| Digits/ Characters | 80% | 76.3% | 20% | 23.7% | 82% | 82.8% | 18% | 17.2% |

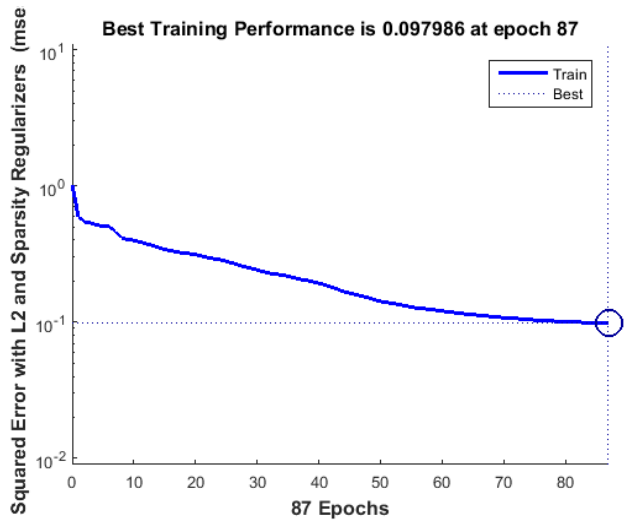

**Fig. 5.** Behavior of error rate reduction for autoeconder training on Digits for 87 epochs. After analysis, it is found that as we increase the number of iterations, the error rate decreases.

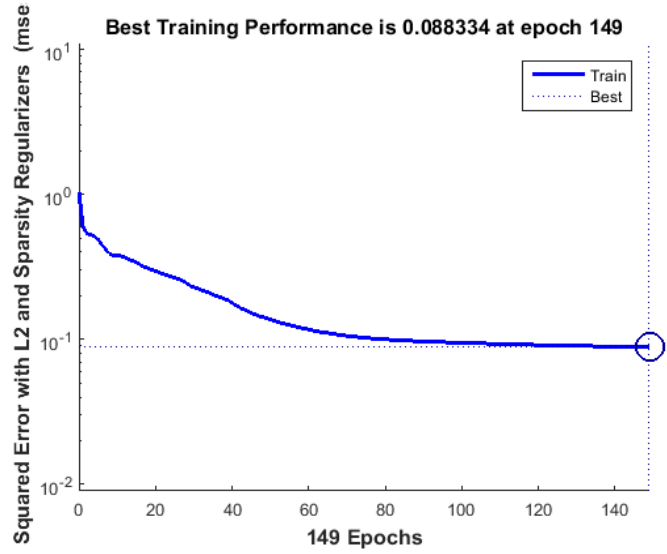

**Fig. 6.** Behavior of error rate reduction for CNN training on Digits for 149 epochs. After analysis, it is found that as we increase the number of iterations, the error rate decreases.

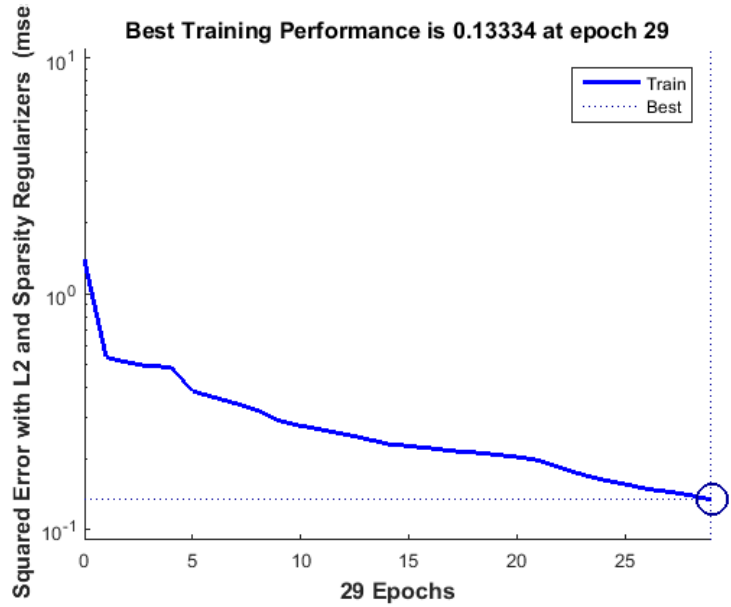

**Fig. 7.** Behavior of error rate reduction for autoeconder training on characters for 29 epochs. After analysis, it is found that as we increase the number of iterations, the error rate decreases.

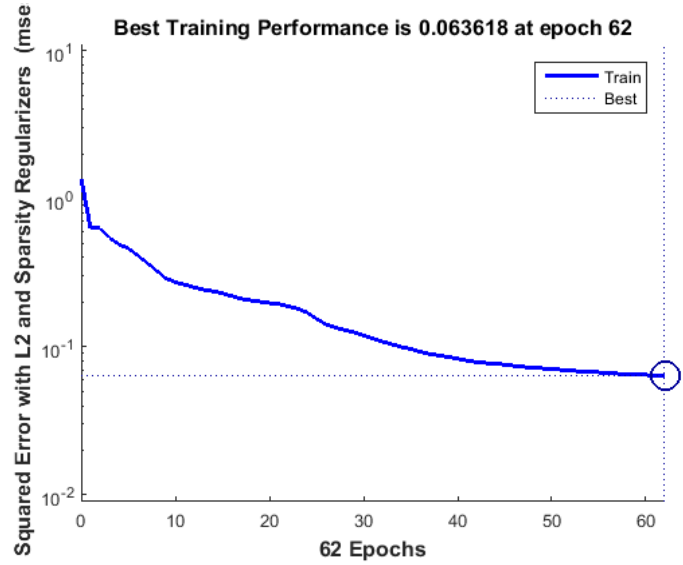

**Fig. 8.** Behavior of error rate reduction for CNN training on characters for 62 epochs. After analysis, it is found that as we increase the number of iterations, the error rate decreases.

### 5.3    Training traditional machine learning classifier

In order to provide an insight into the performance of traditional machine learning classification models, we adopt a heuristic approach with no major optimization and report accuracy of these models for Urdu digits classification. These include results for logistic regression classifier, kNN classifier, a neural network classifier, Gaussian NB, a decision tree model, and an SVM classifier. The accuracy results are reported in Table 6 in the paper.

**Table 6.** Summary of results different traditional classification models

| Sr. No | Name of algorithm | Test Accuracy (%) |
|---|---|---|
| 1 | Logistic regression | 86 |
| 2 | KNN Classifier | 92.09 |
| 3 | Neural Network | 91.98 |
| 4 | Gaussian NB | 69.93 |
| 5 | Decision Tree | 82.00 |
| 6 | SVM | 95.79 |

# 6    Conclusion

In this paper, we have presented a dataset for Urdu hand-written characters and digits, suitable for automatic recognition task. The dataset consists of hand-written samples from 900 individuals. The dataset is randomly divided into train and test sets. The dataset would be made available for free for academic and research use and could be used as baseline as no such dataset is available for Urdu to the best of our knowledge. Further, we have presented a framework for automatic recognition of the hand-written characters and digits. The framework is composed of a three layer autoencoder network and a three layer convolutional neural network trained in a greedy layer-wise fashion. The training has been performed for three different settings of the data namely, for digits only, for characters only and for digits and characters both. For digits recognition, the proposed framework has achieved accuracy up to 97.3% and 96.7%, respectively. For digits and characters, the accuracy is 82% and 82.8%, respectively. Experimental results have shown that using a three layer network results in better recognition performance. These results can act a good baseline for future research and development on Urdu handwritten characters recognition. While it is possible to evaluate many other machine learning algorithms, particularly deep learning algorithms, the use of a variety of algorithms has not been the aim of this paper. Machine learning algorithms such as deep neural networks and generative adversarial networks can be used in future work.

## Acknowledgements

The authors are grateful to Emmanouil Benetos from Queens Mary University London for useful comments on this work.

## Conflict of interest

On behalf of all authors, the corresponding author states that there is no conflict of interest.

**Appendix**

*Appendix-I: Accuracy Rates for digits using 2 hidden layers*

| Digits | Accuracy Rate (%) | Digits | Accuracy Rate (%) |
|--------|-------------------|--------|-------------------|
| ١ | 100 | ٦ | 96.0 |
| ٢ | 94 | ٧ | 97.3 |
| ٣ | 95.2 | ٨ | 99.3 |
| ٤ | 90.3 | ٩ | 99.3 |
| ٥ | 100 | - | - |
| *Total Accuracy Rate* | | *Total Error Rate* | |

*Appendix-II: Accuracy Rates for digits using 3 hidden layers*

| Digits | Accuracy Rate (%) | Digits | Accuracy Rate (%) |
|--------|-------------------|--------|-------------------|
| ١ | 100 | ٦ | 100 |
| ٢ | 92.2 | ٧ | 98.0 |
| ٣ | 96.0 | ٨ | 99.3 |
| ٤ | 92.1 | ٩ | 98.6 |
| ٥ | 100 | - | - |
| *Total Accuracy Rate* | | *Total Error Rate* | |

*Appendix-III: Characters recognition results with two hidden layers – autoencoder framework*

| | Character | Accuracy rate (%) | | Character | Accuracy rate (%) |
|---|---|---|---|---|---|
| Alif | ا | 96.3 | Swad | ص | 61.7 |
| Mad | آ | 86.5 | Zwad | ض | 75 |
| Baa | ب | 79.4 | Twa | ط | 1.7 |
| Paa | پ | 81.5 | Zwaa | ظ | 74.3 |
| Taa | ٹ | 78.1 | Ayn | ع | 80 |
| Tey | ت | 61.3 | Ghain | غ | 71.6 |
| Seey | ث | 65.1 | Faa | ف | 86.1 |
| Jeem | ج | 52.2 | Qaaf | ق | 79.8 |
| Cheey | چ | 72.3 | Kaaf | ک | 76.7 |
| Haa | ح | 67.2 | Gaaf | گ | 83.1 |
| Khaa | خ | 94.9 | Laam | ل | 85.1 |
| Daal | د | 80.3 | Meem | م | 92.7 |
| Zaal | ذ | 78.2 | Noon | ن | 66.1 |
| Dhal | ڈ | 85.9 | Gunna | ں | 71.2 |
| Raa | ر | 75.7 | Wow | و | 84 |
| Zaa | ز | 60 | Haaw | ھ | 76.1 |
| Zaa 2 | ژ | 63.2 | Haaw | ہ | 84 |
| Rhaa | ڑ | 64.3 | Hamza | ء | 80.2 |
| Seen | س | 90 | Choti | ی | 78.8 |
| Sheen | ش | 87.4 | Bari | ے | 96.5 |
| *Total Accuracy Rate for 2 Hidden Layers* | | | *Total error rate for 2 Hidden layers* | | |

*Appendix IV: Characters recognition results for three hidden layers – autoencoder framework*

| | Character | Accuracy rate (%) | | Character | Accuracy rate (%) |
|---|---|---|---|---|---|
| Alif | ا | 93.7 | Swad | ص | 71.8 |
| Mad | ا | 87.5 | Zwad | ض | 86.3 |
| Baa | ب | 82.7 | Twa | ط | 90.2 |
| Paa | پ | 84.9 | Zwaa | ظ | 77 |
| Taa | ٹ | 73.6 | Ayn | ع | 83.6 |
| Tey | ت | 69 | Ghain | غ | 75.8 |
| Seey | ث | 78.4 | Faa | ف | 93.5 |
| Jeem | ج | 53.4 | Qaaf | ق | 80.2 |
| Cheey | چ | 74.3 | Kaaf | ک | 77.4 |
| Haa | ح | 64.5 | Gaaf | گ | 83.9 |
| khaa | خ | 98.3 | Laam | ل | 87.4 |
| Daal | د | 83.2 | Meem | م | 93.8 |
| Zaal | ذ | 81.1 | Noon | ن | 66.7 |
| Dhal | ڈ | 87.9 | Gunna | ں | 79.6 |
| Raa | ر | 84.5 | Wow | و | 82.2 |
| Zaa | ز | 70.1 | Haaw | ھ | 83.8 |
| Zaa 2 | ژ | 68.7 | Haaw | ہ | 88 |
| Rhaa | ڑ | 71.6 | Hamza | ء | 89.1 |
| Seen | س | 94.6 | Choti | ی | 81.6 |
| Sheen | ش | 91.1 | Bari | ے | 92.4 |
| ***Total Accuracy Rate for 3 Hidden Layers*** | | | ***Total error rate for 3 Hidden layers*** | | |

*Appendix V: Digits and characters recognition result for 2 hidden layers – autoencoder framework*

| | Character | Accuracy rate (%) | | Character | Accuracy rate (%) |
|---|---|---|---|---|---|
| 1 | ١ | 61.8 | Zaa 2 | ژ | 63.5 |
| 2 | ٢ | 90.8 | Rhaa | ڑ | 67 |
| 3 | ٣ | 92.1 | Seen | س | 77.1 |
| 4 | ٤ | 88 | Sheen | ش | 86.1 |
| 5 | ٥ | 100 | Swad | ص | 81.8 |
| 6 | ٦ | 94.7 | Zwad | ض | 71.1 |
| 7 | ٧ | 95.4 | Twa | ط | 89.6 |
| 8 | ٨ | 95.8 | Zwaa | ظ | 83.3 |
| 9 | ٩ | 98 | Ayn | ع | 79.8 |
| Alif | ا | 51.8 | Ghain | غ | 81.9 |
| Mad | آ | 91.1 | Faa | ف | 78.4 |
| Baa | ب | 81.4 | Qaaf | ق | 82.6 |
| Paa | پ | 82.8 | Kaaf | ک | 71.5 |
| Taa | ٹ | 76.7 | Gaaf | گ | 69.2 |
| Tey | ت | 64.8 | Laam | ل | 88.3 |
| Seey | ث | 68.8 | Meem | م | 84.3 |
| Jeem | ج | 58.5 | Noon | ن | 92.9 |
| Cheey | چ | 75.5 | Gunna | ں | 73.7 |
| Haa | ح | 62.5 | Wow | و | 80.2 |
| Khaa | خ | 79.5 | Haaw | ھ | 87.6 |
| Daal | د | 79.3 | Haaw | ه | 84.5 |
| Zaal | ڈ | 72.7 | Hamza | ء | 81 |
| Dhal | ذ | 82.3 | Choti | ی | 81.6 |
| Raa | ر | 71.7 | Bari | ے | 96.5 |
| Zaa | ز | 67.4 | - | - | - |
| *Total Accuracy Rate for 2 Hidden Layers* | | | *Total error rate for 2 Hidden layers* | | |

*Appendix VI: Digits and characters recognition result for 3 hidden layers- autoen-coder framework*

| | Character | Accuracy rate (%) | | Character | Accuracy rate (%) |
|---|---|---|---|---|---|
| 1 | ١ | 63.0 | Zaa 2 | ژ | 69.4 |
| 2 | ٢ | 90.9 | Rhaa | ڑ | 72.6 |
| 3 | ٣ | 90.8 | Seen | س | 81.8 |
| 4 | ٤ | 85.3 | Sheen | ش | 94.3 |
| 5 | ٥ | 98 | Swad | ص | 81.8 |
| 6 | ٦ | 97.9 | Zwad | ض | 94.6 |
| 7 | ٧ | 98.6 | Twa | ط | 82.8 |
| 8 | ٨ | 97.1 | Zwaa | ظ | 83.5 |
| 9 | ٩ | 96.7 | Ayn | ع | 85 |
| Alif | ا | 49.2 | Ghain | غ | 70.4 |
| Mad | آ | 89.4 | Faa | ف | 91.8 |
| Baa | ب | 86.5 | Qaaf | ق | 79 |
| Paa | پ | 77.2 | Kaaf | ک | 72.1 |
| Taa | ٹ | 80.8 | Gaaf | گ | 85.8 |
| Tey | ت | 65.4 | Laam | ل | 86.1 |
| Seey | ث | 72.3 | Meem | م | 97.1 |
| Jeem | ج | 60.9 | Noon | ن | 73.9 |
| Cheey | چ | 76.0 | Gunna | ں | 84.2 |
| Haa | ح | 72.2 | Wow | و | 88.9 |
| Khaa | خ | 79.5 | Haaw | ھ | 81.6 |
| Daal | د | 80.7 | Haaw | ہ | 82.6 |
| Zaal | ذ | 80.2 | Hamza | ء | 80.7 |
| Dhal | ڈ | 81.8 | Choti | ی | 84 |
| Raa | ر | 77.5 | Bari | ے | 93.2 |
| Zaa | ز | 69.6 | - | - | - |
| *Total Accuracy Rate for 3 Hidden Layers* | | | *Total error rate for 3 Hidden layers* | | |